\documentclass[11pt,a4paper]{article}
\usepackage[usenames, dvipsnames]{xcolor}
\usepackage[hyperref]{acl2018}
\usepackage{latexsym}
\usepackage[utf8]{inputenc}
\usepackage{amssymb}
\usepackage{booktabs}
\usepackage{graphicx}
\usepackage{times}
\usepackage{enumitem}
\usepackage{multirow}
\usepackage{comment}
\aclfinalcopy
\def\aclpaperid{1637}

\usepackage{xspace}
\newcommand{\blank}{$\underbar{ }\underbar{ }$\xspace}

\title{
Event2Mind: Commonsense Inference on Events, Intents, and Reactions 
}

\author{Hannah Rashkin$^\dagger$\thanks{~~These two authors contributed equally.}~~ Maarten Sap$^\dagger$\footnotemark[1] ~~ Emily Allaway$^\dagger$ ~~ Noah A. Smith$^\dagger$ ~~ Yejin Choi$^\dagger$$^\ddagger$ \\
  $^\dagger$Paul G. Allen School of Computer Science \& Engineering, University of Washington \\
  $^\ddagger$Allen Institute for Artificial Intelligence\\
  {\tt \{hrashkin,msap,eallaway,nasmith,yejin\}@cs.washington.edu} }

\begin{document}

\maketitle

\begin{abstract}
We investigate a new commonsense inference task: given an event described in a short free-form text ({``X drinks coffee in the morning''}), a system reasons about the likely intents ({``X wants to stay awake''}) and reactions ({``X feels alert''}) of the event's participants. 
%
%
To support this study, we construct 
a new crowdsourced corpus of 25,000 event phrases covering a diverse range of everyday events and situations. 
We report baseline performance on this task, demonstrating that neural encoder-decoder models can successfully compose embedding representations of previously unseen events and reason about the likely intents and reactions of the event participants. 
In addition, we demonstrate how commonsense inference on people's intents and reactions can help unveil the implicit gender inequality prevalent in modern movie scripts.

\end{abstract}

\section{Introduction}


\begin{figure}
    \centering
    \includegraphics[width=\columnwidth]{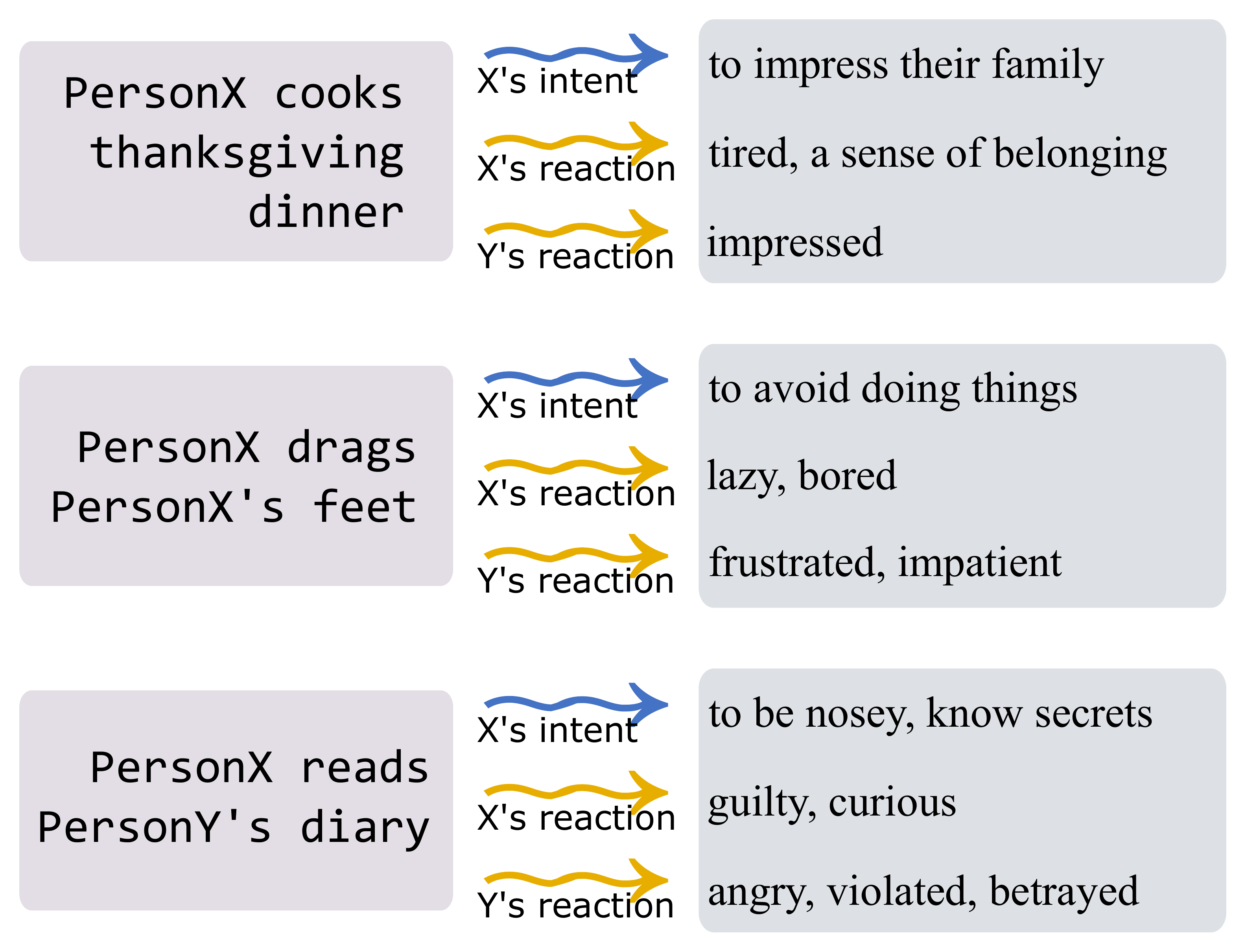}
    \caption{Examples of commonsense inference on mental states of event participants.
    In the third example event, common sense tells us that Y is likely to feel betrayed as a result of X reading their diary.
    }
    \label{fig:commonsenseInferenceExamples}
\end{figure} 

Understanding a narrative  requires 
commonsense reasoning 
about the  mental states of people in relation to events. 
For example, if ``Alex is dragging his feet at work'', pragmatic implications about Alex's \emph{intent} are that ``Alex wants to avoid doing things'' (Figure~\ref{fig:commonsenseInferenceExamples}). We can also infer that Alex's \emph{emotional reaction} might be feeling ``lazy'' or ``bored''. Furthermore, while 
not explicitly mentioned, we can infer that people other than Alex are affected by the situation, and these people are likely to feel ``frustrated'' or ``impatient''. 

This type of pragmatic inference can potentially be useful for a wide range of NLP applications that require accurate anticipation of people's intents and emotional reactions, even when they are not explicitly mentioned. 
%
For example, an ideal dialogue system should react in empathetic ways by reasoning about the human user's mental state based on the events the user has experienced, without the user explicitly stating how they are feeling. 
Similarly, advertisement systems on social media should be able to reason about the emotional reactions of people after events such as mass shootings and remove ads for guns which might increase social distress \cite{FacebookGuns}. Also, pragmatic inference is a necessary step toward automatic narrative understanding and generation \cite{tomai2010using,AffectiveEvents,WhyAffect}.
However, this type of social commonsense reasoning goes far beyond the widely studied entailment tasks \cite{SNLI,dagan2006pascal} and thus falls outside the scope of existing benchmarks.

\begin{table*}[!htb]
\centering

\begin{tabular}{llll}
\toprule
PersonX's Intent  & Event Phrase & PersonX's Reaction & Others' Reactions\\ \midrule
\begin{tabular}[c]{@{}l@{}}  to express anger\\  to vent their frustration \\  to get PersonY's full \\    ~~~attention\end{tabular}                                  & \textbf{\begin{tabular}[c]{@{}l@{}}PersonX starts to \\ yell at PersonY\end{tabular}}    & \begin{tabular}[c]{@{}l@{}}mad\\ frustrated\\ annoyed\end{tabular}         & \begin{tabular}[c]{@{}l@{}}shocked \\ humiliated\\ mad at PersonX\end{tabular} \\ \hline
\begin{tabular}[c]{@{}l@{}} to communicate something \\     ~~~without being rude\\  to let the other person think \\     ~~~for themselves\\  to be subtle\end{tabular} & \textbf{PersonX drops a hint}                                                            & \begin{tabular}[c]{@{}l@{}}sly\\ secretive\\ frustrated\end{tabular}                      & \begin{tabular}[c]{@{}l@{}}oblivious\\ surprised\\ grateful\end{tabular}       \\ \hline
\begin{tabular}[c]{@{}l@{}} to catch the criminal\\  to be civilized\\  justice\end{tabular}                                                        & \textbf{\begin{tabular}[c]{@{}l@{}}PersonX reports \_\_\_ \\ to the police\end{tabular}}    & \begin{tabular}[c]{@{}l@{}}anxious\\ worried\\ nervous \end{tabular} & \begin{tabular}[c]{@{}l@{}}sad\\ angry\\ regret \end{tabular}                                                                     \\ \hline
\begin{tabular}[c]{@{}l@{}} to wake up\\  to feel more energized\end{tabular}                                                                                     & \textbf{\begin{tabular}[c]{@{}l@{}}PersonX drinks \\ a cup of coffee\end{tabular}}       & \begin{tabular}[c]{@{}l@{}}alert\\ awake\\ refreshed\end{tabular}            & NONE                                                                           \\ \hline
\begin{tabular}[c]{@{}l@{}} to be feared\\  to be taken seriously\\  to exact revenge\end{tabular}                                                                 & \textbf{\begin{tabular}[c]{@{}l@{}}PersonX carries \\ out PersonX's threat\end{tabular}} & \begin{tabular}[c]{@{}l@{}}angry\\ dangerous\\ satisfied\end{tabular}        & \begin{tabular}[c]{@{}l@{}}sad\\ afraid \\ angry\end{tabular} 
                                           \\ \hline
NONE                                                  & \textbf{\begin{tabular}[c]{@{}l@{}} It starts \\ snowing \end{tabular}} & NONE        & \begin{tabular}[c]{@{}l@{}}calm\\ peaceful \\ cold \end{tabular} \\ \bottomrule
\end{tabular}
\caption{Example annotations of intent and reactions for 6 event phrases.
Each annotator could fill in up to three free-responses for each mental state.}
\label{tab:annotationex}
\end{table*}

In this paper, we introduce a new task, corpus, and model, supporting commonsense inference on events with a specific focus on modeling stereotypical intents and reactions of people, described in short free-form text. 
Our study is in a similar spirit to recent efforts of \citet{AffectiveEvents} and \citet{Zhang2017OrdinalCI}, in that we aim to model aspects of commonsense inference via natural language descriptions. 
Our new contributions are: (1) a new  corpus that supports commonsense inference about people's intents and reactions over a diverse range of everyday events and situations, (2) inference about even those people who are not directly mentioned by the event phrase, and (3) a task formulation that aims to \emph{generate} the textual descriptions of intents and reactions, instead of classifying their polarities or
classifying the inference relations between two given textual descriptions.


%

Our work establishes baseline performance on this new task, demonstrating that, given the phrase-level inference dataset, neural encoder-decoder models can successfully compose phrasal embeddings for previously unseen events and reason about the mental states of their participants. 

Furthermore, in order to showcase the practical implications of commonsense inference on events and people's mental states, 
we apply our model to modern movie scripts, which provide a new insight into the gender bias in modern films beyond what previous studies have offered \cite{england2011gender,Agarwal2015-lq,Ramakrishna2017,Sap2017-lt}.
The resulting corpus includes around 25,000 event phrases, which combine automatically extracted phrases from stories and blogs with all idiomatic verb phrases listed in the Wiktionary. Our corpus is publicly available.\footnote{\url{https://tinyurl.com/event2mind}}

\section{
Dataset}

One goal of our investigation is to probe whether it is feasible to build computational models that can perform limited, but well-scoped commonsense inference on short free-form text, which we refer to as \emph{event phrases}. While there has been much prior research on phrase-level paraphrases \cite{Pavlick2015PPDB2B} and phrase-level entailment \cite{dagan2006pascal}, relatively little prior work focused on phrase-level inference that requires pragmatic or commonsense interpretation. We scope our study to two distinct types of inference: given a phrase that describes an event, we want to reason about the likely intents and emotional reactions of people who caused or affected by the event. This complements prior work on more general commonsense inference \cite{ConceptNet,Li2016CommonsenseKB,Zhang2017OrdinalCI}, by focusing on the causal relations between events and people's mental states, which are not well covered by most existing resources.


We collect a wide range of phrasal event descriptions from stories, blogs, and Wiktionary  idioms. Compared to prior work on phrasal embeddings \cite{Wieting2015FromPD,Pavlick2015PPDB2B}, our work  generalizes the phrases by introducing (typed) variables. In particular, we replace words that correspond to entity mentions or pronouns with typed variables such as \texttt{PersonX} or \texttt{PersonY}, as shown in examples in Table~\ref{tab:annotationex}. More formally, the phrases we extract are a combination of a verb predicate with partially instantiated arguments. We keep specific arguments   together with the predicate, if they appear frequently enough (e.g., \texttt{PersonX eats pasta for dinner}). Otherwise, the arguments are replaced with an untyped blank (e.g., \texttt{PersonX eats \blank for dinner}). In our work, only person mentions are replaced with typed variables, leaving other types to future research.


\paragraph{Inference types} 
The first type of pragmatic inference is about \textit{intent}.
We define intent as an explanation of why the agent causes a volitional event to occur (or ``none'' if the event phrase was unintentional). The intent can be considered a mental pre-condition of an action or an event.  For example, if the event phrase is \texttt{PersonX takes a stab at \blank}, the annotated intent might be that ``PersonX wants to solve a problem''.

The second type of pragmatic inference is about \textit{emotional reaction}. 
We define reaction as an explanation of how the mental states of the agent and other people involved in the event would change as a result. The reaction can be considered a mental post-condition of an action or an event.  For example, if the event phrase is that \texttt{PersonX gives PersonY \blank as a gift}, PersonX might ``feel good about themselves'' as a result, and PersonY might ``feel grateful'' or ``feel thankful''.

\begin{table}[tb]
\centering
\small
\begin{tabular}{lrrrr} 
\toprule
Source   & \begin{tabular}[c]{@{}r@{}}\# Unique \\ Events\end{tabular} & \begin{tabular}[c]{@{}r@{}}\# Unique \\ Verbs\end{tabular} &  \begin{tabular}[c]{@{}r@{}}Average \\  $\kappa$\end{tabular}\\\midrule 
ROC Story & 13,627           & 639  & 0.57\\  
G.~N-grams  & 7,066            & 789 & 0.39\\ 
Spinn3r  & 2,130            & 388  & 0.41 \\          
Idioms   & 1,916            & 442 & 0.42 \\\midrule  
\textbf{Total}    & \textbf{24,716}          & \textbf{1,333} & \textbf{0.45}\\\bottomrule
\end{tabular}
\caption{Data and annotation agreement statistics for our new phrasal inference corpus. Each event is annotated by three crowdworkers.} 
\label{tab:distrib}
\end{table}

\subsection{Event Extraction}
We extract phrasal events from three different corpora for broad coverage:
the ROC Story training set \cite{Mostafazadeh2016-ei}, the Google Syntactic N-grams \cite{Ngrams}, and the Spinn3r corpus \cite{Spinn3r}.
We derive events from the set of verb phrases in our corpora, based on syntactic parses  \cite{klein2003accurate}.
We then replace the predicate subject and other entities with the typed variables (e.g., \texttt{PersonX}, \texttt{PersonY}), and selectively substitute verb arguments with blanks (\texttt{\blank}).
We use frequency thresholds to select events to annotate (for details, see Appendix~\ref{ss:event_extraction_more}).
Additionally, we supplement the list of events with all 2,000 verb idioms found in Wiktionary, in order to cover events that are less compositional.\footnote{We compiled the list of idiomatic verb phrases by cross-referencing between Wiktionary's English idioms category and the Wiktionary English verbs categories.}
Our final annotation corpus contains nearly 25,000 event phrases, spanning over 1,300 unique verb predicates (Table~\ref{tab:distrib}).

\subsection{Crowdsourcing}
\begin{figure}[t]
    \centering
    \fbox{\includegraphics[width=.95\columnwidth,trim={0 1cm 0 0},clip]{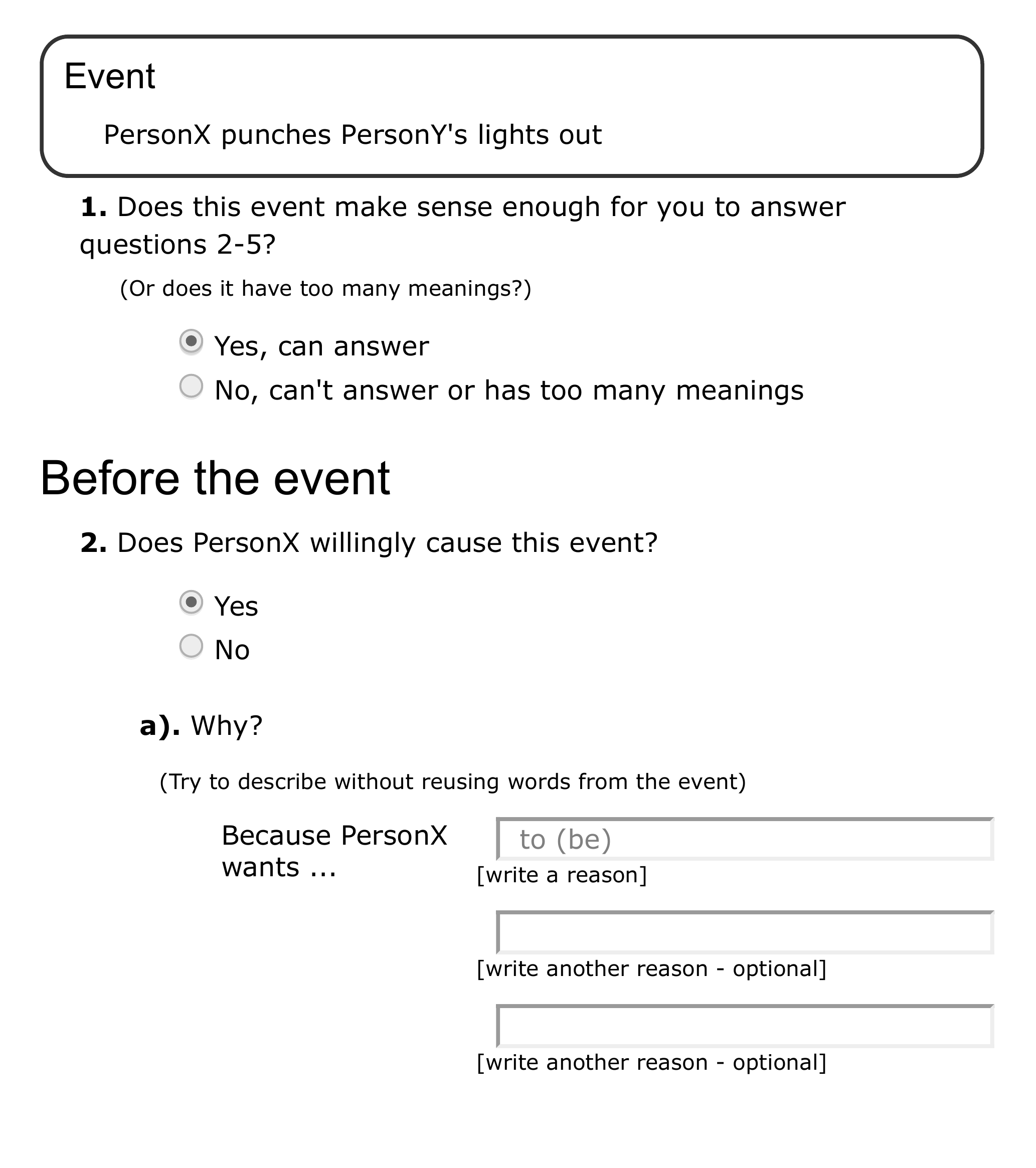}}
    \caption{\textit{Intent} portion of our annotation task. We allow annotators to label events as invalid if the phrase is unintelligible. The full annotation setup is shown in Figure~\ref{fig:mturkSetup} in the appendix.
    \label{fig:mturkSetupShort}}
\end{figure}
We design an Amazon Mechanical Turk task to annotate the mental pre- and post-conditions of event phrases. A snippet of our MTurk HIT design is shown in Figure~\ref{fig:mturkSetupShort}.
For each phrase, we ask three annotators whether the agent of the event, PersonX, intentionally causes the event, and if so, to provide up to three possible textual descriptions of their intents.
We then ask annotators to provide up to three possible reactions that PersonX might experience as a result. 
We also ask annotators to provide up to three possible reactions of \emph{other people}, when applicable. These other people can be either explicitly mentioned (e.g., ``PersonY'' in \texttt{PersonX punches PersonY's lights out}), or only implied (e.g., given the event description \texttt{PersonX yells at the classroom}, we can infer that other people such as ``students'' in the classroom may be affected by the act of PersonX). 
For quality control, 
we periodically removed workers with high disagreement rates, at our discretion.


%

\paragraph{Coreference among \texttt{Person} variables} 
With the typed \texttt{Person} variable setup, events involving multiple people can have multiple meanings depending on coreference interpretation (e.g., \texttt{PersonX eats PersonY's lunch} has very different mental state implications from \texttt{PersonX eats PersonX's lunch}).
To prune the set of events that will be annotated for intent and reaction, we ran a preliminary annotation to filter out candidate events that have implausible coreferences.
In this preliminary task, annotators were shown a combinatorial list of coreferences for an event (e.g., \texttt{PersonX punches PersonX's lights out}, \texttt{PersonX punches PersonY's lights out}) and were asked to select only the plausible ones (e.g., \texttt{PersonX punches PersonY's lights out}).
Each set of coreferences was annotated by 3  workers, yielding an overall agreement of  $\kappa=$0.4.
This annotation excluded 8,406 events with implausible coreference from our set (out of 17,806 events).

\subsection{Mental State Descriptions}
Our  dataset contains nearly 25,000 event phrases, with annotators rating 91\% of our extracted events as ``valid'' (i.e., the event makes sense).
Of those events, annotations for the multiple choice portions of the task (whether or not there exists intent/reaction) agree moderately, with an average Cohen's $\kappa=$ 0.45 (Table~\ref{tab:distrib}).
The individual $\kappa$ scores generally indicate that turkers disagree half as often as if they were randomly selecting answers. 

Importantly, this level of agreement is acceptable in our task formulation for two reasons. First, 
unlike linguistic annotations on syntax or semantics where experts in the corresponding theory would generally agree on a single correct label, pragmatic interpretations may better be defined as distributions over multiple correct labels \citep[e.g., after \texttt{PersonX takes a test}, PersonX might feel relieved and/or stressed;][]{Marneffe2012DidIH}.
Second, because we formulate our task as a conditional language modeling problem, where  a distribution over the textual descriptions of intents and reactions is conditioned on the event description, this variation in the labels is only as expected.

A majority of our events are annotated as willingly caused by the agent (86\%,
 Cohen's $\kappa=$ 0.48), and 26\% involve other people ($\kappa=$ 0.41).
Most event patterns in our data are fully instantiated, with only 22\% containing blanks (\texttt{\blank}).
In our corpus, the intent annotations are slightly longer (3.4 words on average) than the reaction annotations (1.5 words).

\section{Models}
\begin{figure}[tb]
\centering
\includegraphics[width=.45\textwidth,page=1]{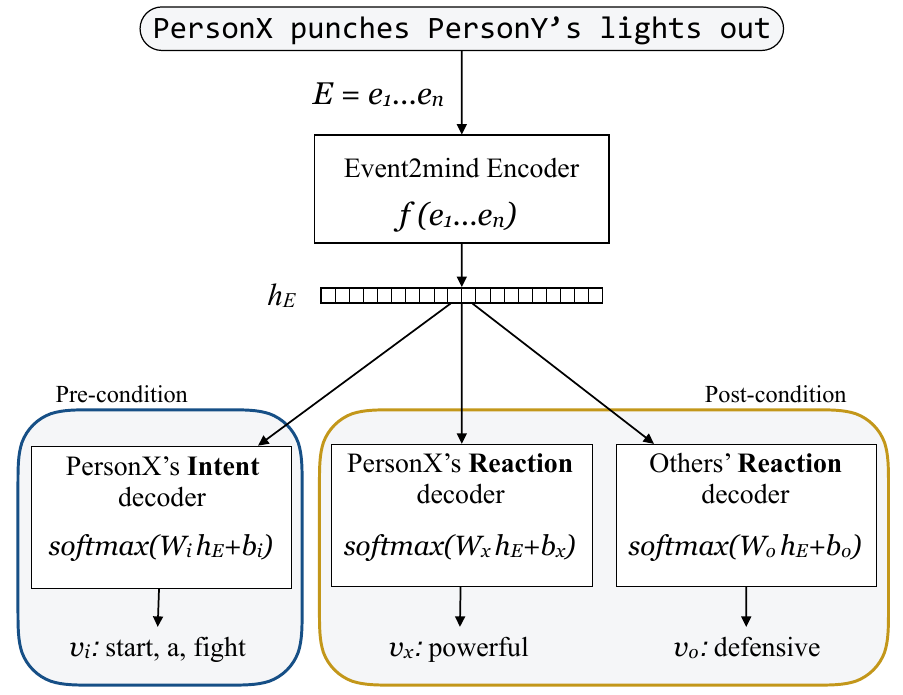}
\caption{Overview of the model architecture. From an encoded event, our model predicts intents and reactions in a multitask setting.}
\label{fig:modelArchitecture}
\end{figure}

Given an event phrase, our models aim to generate three entity-specific pragmatic inferences:  PersonX's intent, PersonX's reaction, and others' reactions. 
The general outline of our model architecture is illustrated in Figure~\ref{fig:modelArchitecture}.


The input to our model is an event pattern described through free-form text with typed variables such as \texttt{PersonX gives PersonY \_\_\_ as a gift}.
For notation purposes, we describe each event pattern $E$ as a sequence of word embeddings $\langle e_1, e_2, \ldots, e_n\rangle \in \mathbb{R}^{n\times D}$.  This input is encoded as a vector $h_E \in \mathbb{R}^H$ that will be used for predicting output.
The output of the model is its hypotheses about PersonX's intent, PersonX's reaction, and others' reactions ($v_i$,$ v_x$, and $v_o$, respectively).  We experiment with representing the output in two decoding set-ups: three vectors interpretable as discrete distributions over words and phrases (n-gram reranking) or three sequences of words (sequence decoding).  
%

\paragraph{Encoding events}
The input event phrase $E$ is compressed into an $H$-dimensional embedding $h_E$ via an encoding function $f: \mathbb{R}^{n\times D} \rightarrow \mathbb{R}^{H}$:
\[h_E = f(e_1, \ldots, e_n)\]

\noindent
We experiment with several ways for defining $f$, inspired by standard techniques in sentence and phrase classification~\citep{YoonKimPaper}. 
First, we experiment with max-pooling and mean-pooling over the word vectors $\{e_i\}_{i=1}^n$. 
We also consider a convolutional neural network \citep[ConvNet;][]{lecun1998gradient}
taking the last layer of the network as the encoded version of the event.
Lastly, we encode the event phrase with a bi-directional RNN \citep[specifically, a GRU; ][]{Cho2014}, concatenating the final hidden states of the forward and backward cells as the encoding: $h_E = [\overrightarrow{h_n};\overleftarrow{h_1}]$. For hyperparameters and other details, we refer the reader to Appendix~\ref{sec:trainingdeets}.

Though the event sequences are typically rather short (4.6 tokens on average), our model still benefits from the ConvNet and BiRNN's ability to compose words. 

\paragraph{Pragmatic inference decoding}
We use three decoding modules that take the event phrase embedding $h_E$ and output distributions of possible PersonX's intent ($v_i$), PersonX's reactions ($v_x$), and others' reactions ($v_o$). We experiment with two different decoder set-ups.

%
First, we experiment with \textit{n-gram re-ranking}, considering the $|V|$ most frequent \{1, 2, 3\}-grams in our annotations.
Each decoder projects the event phrase embedding $h_E$ into a $|V|$-dimensional vector, which is then passed through a softmax function. For instance, 
the distribution over descriptions of PersonX's intent is given by:
\[v_i = \textrm{softmax}(W_i h_E +b_i) \] 


\noindent Second, we experiment with \textit{sequence generation}, using RNN decoders to generate the textual description.
The event phrase embedding $h_E$ is set as the initial state $h_{dec}$ of three decoder RNNs (using GRU cells), which then output the intent/reactions one word at a time (using beam-search at test time).
For example, an event's intent sequence $(v_i = v_i^{(0)}v_i^{(1)}\ldots)$ is computed as follows: 
\[v_i^{(t+1)} = \textrm{softmax}(W_i~\textrm{RNN}(v_i^{(t)}, h_{i,dec}^{(t)}) + b_i) \]

\paragraph{Training objective}
We minimize the cross-entropy between the predicted distribution over words and phrases, against the one actually observed in our dataset. 
Further, we employ multitask learning, simultaneously minimizing the loss for all three decoders at each iteration.

\paragraph{Training details}
We fix our input embeddings, using 300-dimensional skip-gram 
word embeddings trained on Google News~\cite{Mikolov2013EfficientEO}. 
For decoding, we consider a vocabulary of size $|V|=$ 14,034 in the n-gram re-ranking setup.
For the sequence decoding setup, we only consider the unigrams in $V$, yielding an output space of 7,110 at each time step.

We randomly divided our set of 24,716 unique events (57,094 annotations) into a training/dev./test set using an 80/10/10\% split.  Some annotations have multiple responses (i.e., a crowdworker gave multiple possible intents and reactions), in which case we take each of the combinations of their responses as a separate training example. 


\begin{table*}[tb]
\centering
\begin{tabular}{@{}l@{\hspace{10pt}}l@{\hspace{5pt}}c@{\hspace{7pt}}r@{\hspace{7pt}}r@{\hspace{7pt}}r@{}@{\hspace{10pt}}r@{}@{\hspace{10pt}}c@{\hspace{7pt}}r@{\hspace{7pt}}r@{\hspace{7pt}}r@{}}
\\ & & \multicolumn{4}{c}{Development}&&\multicolumn{4}{c}{Test}\\
\cmidrule{3-6}\cmidrule{8-11}
\multirow{2}{*}{\begin{tabular}[c]{@{}l@{}}Encoding\\ Function\end{tabular}} &
\multirow{2}{*}{\begin{tabular}[c]{@{}l@{}}Decoder\end{tabular}} & \multirow{2}{*}{\begin{tabular}[c]{@{}c@{}}Average\\ Cross-Ent\end{tabular}} & \multicolumn{3}{c}{Recall @10 (\%)}& & \multirow{2}{*}{\begin{tabular}[c]{@{}c@{}}Average\\ Cross-Ent\end{tabular}} & \multicolumn{3}{c}{Recall @10 (\%)} \\
 &  & &  Intent & XReact & OReact && & Intent & XReact & OReact \\ \midrule
max-pool & n-gram& 5.75 & 31 & 35 & 68  && 5.14 & 31 & 37 & 67\\
mean-pool & n-gram& 4.82 & 35 & 39 & 69 && 4.94 & 34 & 40 & 68\\
ConvNet &n-gram & 4.85 & 36 & 42 & 69 && 4.81 & 37 & 44 & 69\\
BiRNN 300d & n-gram&  4.78 & 36 & 42 & 68 && 4.74 & 36 & 43& 69\\
BiRNN 100d & n-gram &  4.76 & 36 & 41 & 68 && 4.73 & 37 & 43 & 68 \\
\midrule
mean-pool& sequence &   4.59&  39 & 36& 67 && 4.54 &40 & 38& 66 \\ 
ConvNet & sequence &  4.44& 42 & 39 &  68 && 4.40& 43 &40 & 67 \\ 
BiRNN 100d & sequence &  4.25 & 39 & 38 & 67 && 4.22 &40 & 40 & 67\\ 
\bottomrule
\end{tabular}
\caption{Average cross-entropy (lower is better) and recall @10 (percentage of times the gold falls within the top 10 decoded; higher is better) on development and test sets for different modeling variations.
We show recall values for PersonX's intent, PersonX's reaction and others' reaction (denoted as ``Intent'', ``XReact'', and ``OReact'').
Note that because of two different decoding setups, cross-entropy between n-gram and sequence decoding are not directly comparable.}
\label{dev:intrinsic}
\end{table*}



\section{Empirical Results}


Table~\ref{dev:intrinsic} summarizes the performance of different encoding models on the dev and test set in terms of cross-entropy
and recall at 10 predicted intents and reactions.
As expected, we see a moderate improvement in recall and cross-entropy when using the more compositional encoder models (ConvNet and BiRNN; both n-gram and sequence decoding setups). 
Additionally, BiRNN models outperform ConvNets on cross-entropy in both decoding setups.
Looking at the recall split across intent vs.~reaction labels (``Intent'', ``XReact'' and ``OReact'' columns), we see that much of the improvement in using these two models is within the prediction of PersonX's intents.
Note that recall for ``OReact'' is much higher, since a majority of events do not involve other people.

\paragraph{Human evaluation}
To further assess the quality of our models, we randomly select 100 events from our test set and ask crowd-workers to rate generated intents and reactions.
We present 5 workers with an event's top 10 most likely intents and reactions according to our model and ask them to select all those that make sense to them.
We evaluate each model's precision @10 by computing the average number of generated responses that make sense to annotators.

\begin{figure}[th]
\centering
\includegraphics[width=\columnwidth,trim=2cm 2.5cm 4.5cm 2cm]{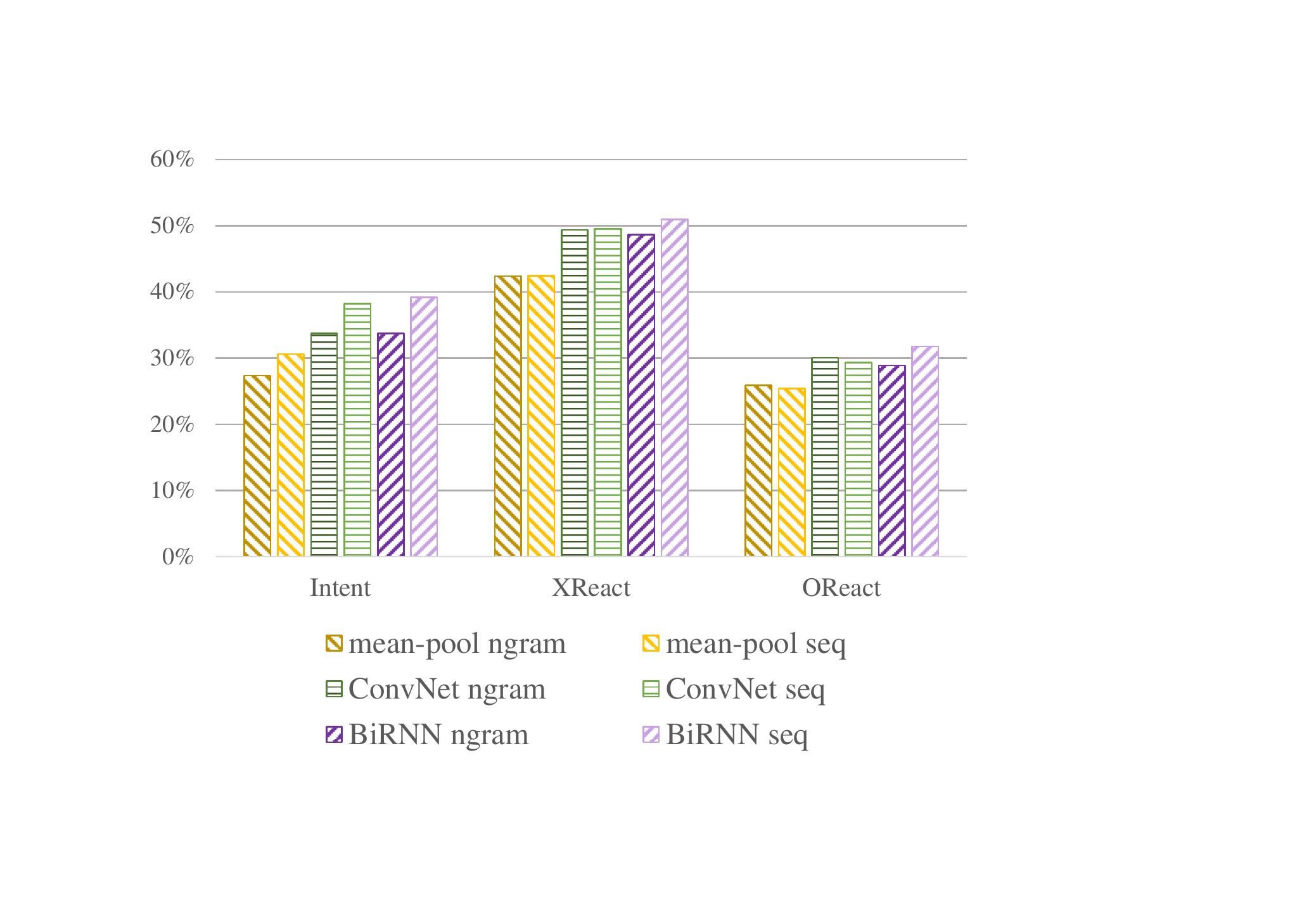}
\caption{Average precision @10 of each model's top ten responses in the human evaluation. We show results for various encoder functions (mean-pool, ConvNet, BiRNN-100d) combined with two decoding setups (n-gram re-ranking, sequence generation).}
\label{fig:humaneval}
\end{figure}

Figure~\ref{fig:humaneval} summarizes the results of this evaluation. In most cases, the performance is higher for the sequential decoder than the corresponding n-gram decoder.  The biggest gain from using sequence decoders is in intent prediction, possibly because intent explanations are more likely to be longer. The BiRNN and ConvNet encoders consistently have higher precision than the mean-pooling with the BiRNN-seq setup slightly outperforming other models.
Unless otherwise specified, this is the model we employ in further sections.

\paragraph{Error analyses}

\begin{figure*}[tb]
    \centering
    \includegraphics[width=1.0\textwidth]{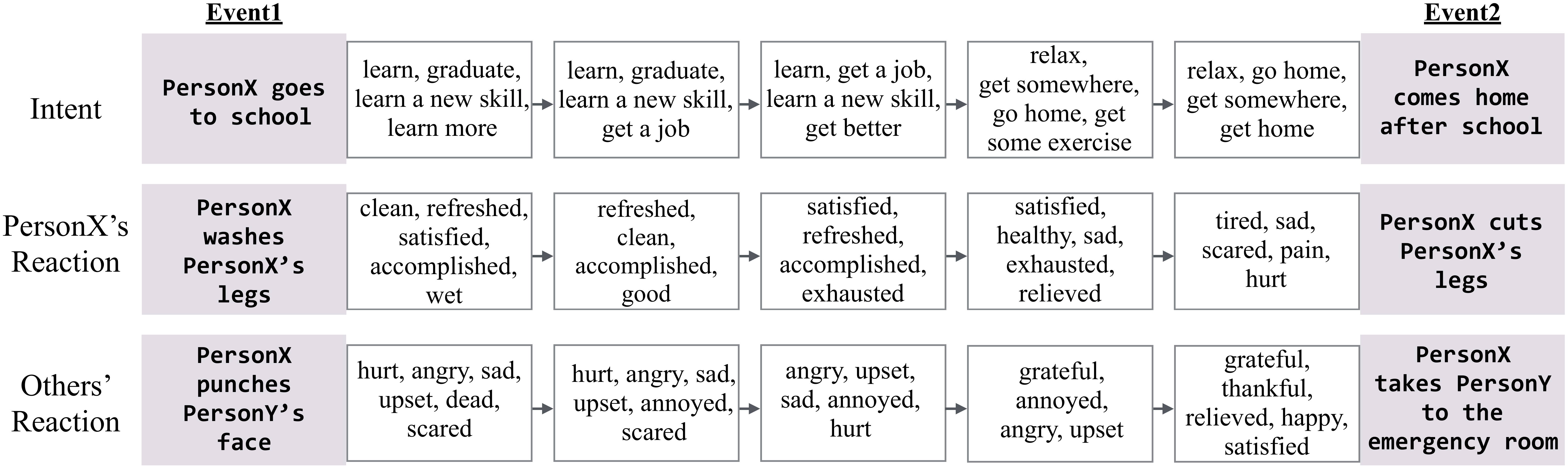}
    \caption{Sample predictions from homotopic embeddings (gradual interpolation between Event1 and Event2), selected from the top 10 beam elements decoded in the sequence generation setup. Examples highlight differences captured when ideas are similar (\textit{going to} and \textit{coming from} school), when only a single word differs (\textit{washes} versus \textit{cuts}), and when two events are unrelated.} 
    \label{fig:homotopies}
\end{figure*}

\begin{figure}[t]
    \centering
    \includegraphics[width=1\columnwidth,trim=0 0 0 20pt]{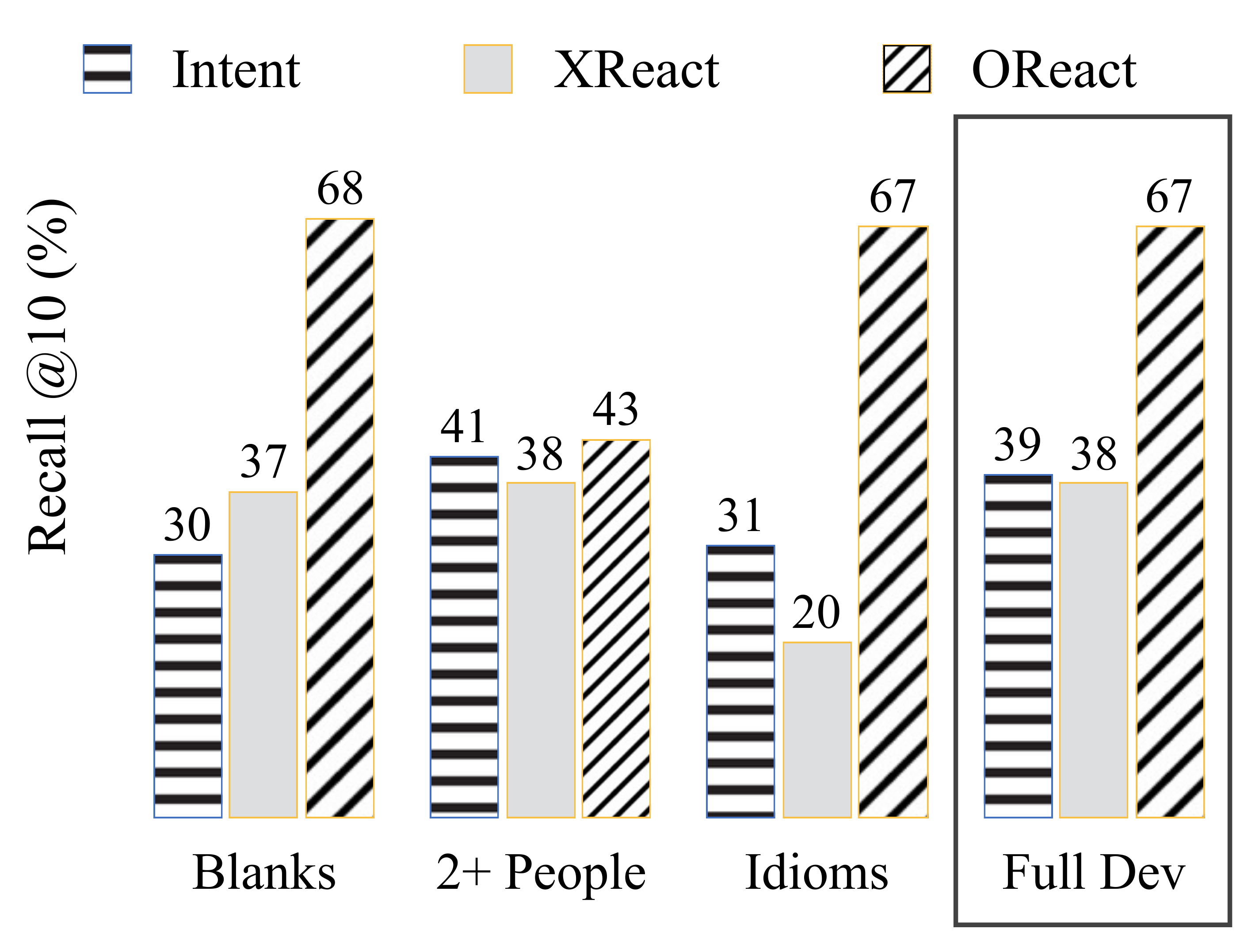}
    \caption{Recall @ 10 (\%) on different subsets of the development set for intents, PersonX's reactions, and other people's reactions, using the BiRNN 100d model. ``Full dev'' represents the recall on the entire development dataset.}
    \label{fig:ablations}
\end{figure}




We test whether certain types of events 
are easier for predicting commonsense inference.
In Figure~\ref{fig:ablations}, we show the difference in cross-entropy of the BiRNN 100d model on predicting different portions of the development set including:
\texttt{Blank} events (events containing non-instantiated arguments), \texttt{2+ People} events (events containing multiple different Person variables), and \texttt{Idiom} events (events coming from the Wiktionary idiom list). 
Our results show that, while intent prediction performance remains similar for all three sets of events, it is 10\% behind intent prediction on the full development set.
Additionally, predicting other people's reactions is more difficult for the model when 
other people are explicitly mentioned. 
Unsurprisingly, idioms are particularly difficult for commonsense inference, perhaps due to the difficulty in composing meaning over nonliteral or noncompositional event descriptions.

To further evaluate the geometry of the embedding space, we analyze interpolations between pairs of event phrases (from outside the train set), similar to the homotopic analysis of \citet{Bowman2016}.  For a handful of event pairs, we decode intents, reactions for PersonX, and reactions for other people from points sampled at equal intervals on the interpolated line between two event phrases. We show examples in Figure~\ref{fig:homotopies}. The embedding space distinguishes changes from generally positive to generally negative words and is also able to capture small differences between event phrases (such as ``washes'' versus ``cuts'').
\section{Analyzing Bias via Event2Mind Inference}

\begin{figure}[ht]
    \centering
    \includegraphics[width=\columnwidth]{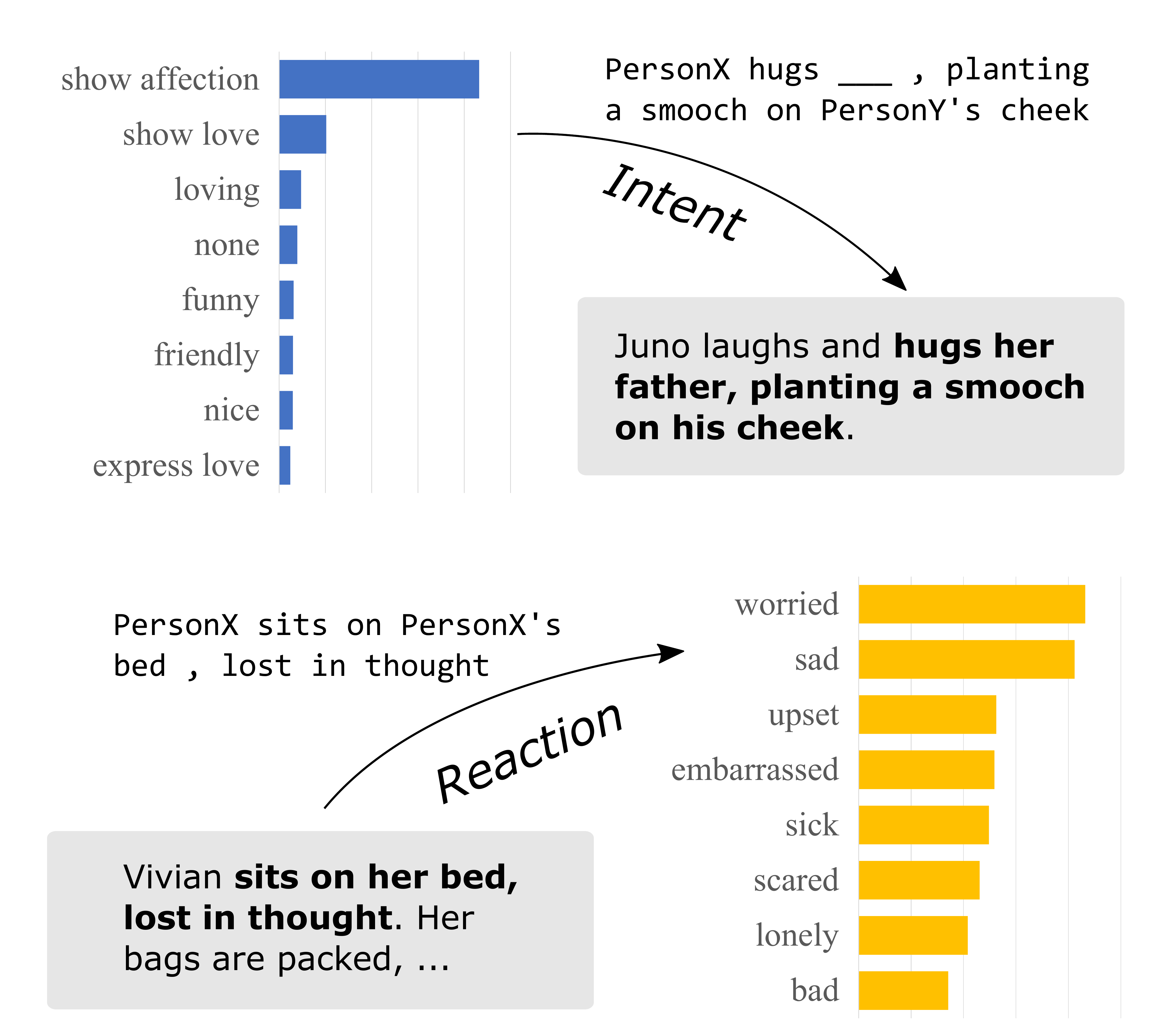}
    \caption{
    Two scene description snippets from \textit{Juno} (2007, top) and \textit{Pretty Woman} (1990, bottom), augmented with Event2mind inferences on the characters' intents and reactions.
    E.g., our model infers that the event \texttt{PersonX sits on PersonX's bed, lost in thought} implies that the agent, Vivian, is sad or worried.}
    \label{fig:moviesQual}
\end{figure}

Through Event2Mind inference, 
we can attempt to bring to the surface what is implied about people's behavior and mental states.
We employ this inference to analyze implicit bias in
modern films.
%
As shown in Figure \ref{fig:moviesQual}, our model is able to analyze character portrayal beyond what is explicit in text, by performing pragmatic inference on character actions to explain aspects of a character's mental state.
In this section, we use our model's inference to shed light on gender differences in intents behind and reactions to characters' actions. 

\subsection{Processing of Movie Scripts}
For our portrayal analyses, we use scene descriptions from 772 movie scripts released by \citet{Gorinski2015-no}, assigned to over 21,000 characters as done by \citet{Sap2017-lt}.
We extract events from the scene descriptions, and generate their 10 most probable intent and reaction sequences using our BiRNN sequence model (as in Figure~\ref{fig:moviesQual}). 

We then categorize generated intents and reactions into groups based on LIWC category
scores  of the generated output \cite{Tausczik2016-rd}.\footnote{We only consider content word categories: 
`Core Drives and Needs', `Personal Concerns', `Biological Processes', `Cognitive Processes', `Social Words', `Affect Words', `Perceptual Processes'.
We refer the reader to \citet{Tausczik2016-rd} or \url{http://liwc.wpengine.com/compare-dictionaries/} for a complete list of category descriptions.
}
The intent and reaction categories are then aggregated for each character, and 
standardized (zero-mean and unit variance).

We compute correlations with gender for each category of intent or reaction using a logistic regression model, testing significance while using Holm's correction for multiple comparisons~\cite{holm1979simple}.\footnote{Given the data limitation, we represent gender as a binary, but acknowledge that gender is a more complex social construct.}
To account for the gender skew in scene presence (29.4\% of scenes have women), we statistically control for the total number of words in a character's scene descriptions.
Note that the original event phrases are all gender agnostic, as their participants have been replaced by variables (e.g., \texttt{PersonX}). We also find that the types of gender biases uncovered remain similar when we run these analyses on the human annotations or the generated words and phrases from the BiRNN with n-gram re-ranking decoding setup.


\subsection{Revealing Implicit Bias via Explicit Intents and Reactions}

\begin{table}[ht]
\centering


\begin{tabular}{p{.9\columnwidth}}
\toprule
\multicolumn{1}{c}{\textbf{Female: intents}}         \\
\textsc{affiliation, friend, family} \\
\textsc{body, sexual, ingest}\\
\textsc{see, insight, discrep}\\
\multicolumn{1}{c}{\textbf{Male: intents}}           \\
\textsc{death, health, anger, negemo}                        \\
\textsc{risk, power, achieve, reward, work}       \\
\textsc{cause, tentative$^\ddagger$}       \\
\midrule
\multicolumn{1}{c}{\textbf{Female: reactions}}       \\ 
\textsc{posemo, affiliation, friend, reward}   \\
\textsc{ingest, sexual$^\ddagger$, body$^\ddagger$}                   \\
\multicolumn{1}{c}{\textbf{Male: reactions}}         \\ 
\textsc{work, achieve, power, health$^\dagger$}                        \\
\midrule
\multicolumn{1}{c}{\textbf{Female: others' reactions}}\\ 
\textsc{posemo, affiliation, friend}  \\
\textsc{ingest, see, insight}\\
\multicolumn{1}{c}{\textbf{Male: others' reactions}}\\ 
\textsc{achieve, risk$^\dagger$} \\
\textsc{sad, negemo$^\ddagger$, anger$^\dagger$} \\
\bottomrule
\end{tabular}

\caption{Select LIWC categories correlated with gender. All results are significant when corrected for multiple comparisons at $p<0.001$, except $^\dagger p<0.05$ and $^\ddagger p<0.01$.
}
\label{tab:LIWCmovies}
\end{table}

Our Event2Mind inferences automate portrayal analyses that previously required manual annotations \cite{behm2008mean,prentice2002women,england2011gender}.
Shown in Table~\ref{tab:LIWCmovies}
,
our results indicate a  gender bias in the behavior ascribed to characters, consistent with psychology and gender studies literature \cite{collins2011content}. 
Specifically, events with female semantic agents are intended to be helpful to other people (intents involving \textsc{friend}, \textsc{family}, and \textsc{affiliation}), particularly relating to eating and making food for themselves and others (\textsc{ingest}, \textsc{body}).
Events with male agents on the other hand are motivated by and resulting in achievements (\textsc{achieve}, \textsc{money}, \textsc{rewards}, \textsc{power}).

Women's looks and sexuality are also 
emphasized, as their actions' intents and reactions are sexual, seen, or felt (\textsc{sexual}, \textsc{see}, \textsc{percept}).
Men's actions, on the other hand, are motivated by violence or fighting (\textsc{death}, \textsc{anger}, \textsc{risk}), with strong negative reactions (\textsc{sad}, \textsc{anger}, \textsc{negative emotion}).

Our approach decodes nuanced implications into  more explicit statements, helping to  identify and explain  gender bias that is  prevalent in modern literature and media. 
Specifically, our results indicate that modern movies have the bias to portray female characters as having  pro-social attitudes, whereas male characters are portrayed as being competitive or pro-achievement.
This is consistent with gender stereotypes that have been studied in movies in both NLP and psychology literature \cite{Agarwal2015-lq,Madaan2017AnalyzingGS,prentice2002women,england2011gender}.

\section{Related Work}

Prior work has sought formal frameworks for inferring roles and other attributes in relation to events \cite[][\emph{inter alia}]{FrameNet,das:2014:cl,VerbNet, VerbCorner}, implicitly connoted by events \cite{SemanticProtoroles, UniversalDecomp, ImplicitSent, Rashkin2016}, or   
%
sentiment polarities of events 
\cite{AffectiveEvents,gfbfpaper,SemEval2015,ding2018weakly}. In addition, recent work has studied the patterns which evoke certain polarities \cite{LexicoFP}, the desires which make events affective \cite{WhyAffect}, the emotions caused by events \cite{Vu2014}, or, conversely, identifying events or reasoning behind particular emotions \cite{EmotionCause}.  
Compared to this prior literature, 
our work uniquely learns to model intents and reactions over a diverse set of events, includes inference over event participants not explicitly mentioned in text, and formulates the task as predicting the textual descriptions of the implied commonsense instead of classifying various event attributes.


Previous work in natural language inference has focused on linguistic entailment \cite{SNLI,Bos2005} while ours  focuses on commonsense-based inference. There also has been inference or entailment work that is more generation focused: generating, e.g.,  entailed statements \cite{Zhang2017OrdinalCI,InferenceRoles}, explanations of causality \cite{Kang2017}, or  paraphrases  \cite{Dong2017LearningTP}.  
Our work also aims at generating inferences from sentences; however, our models infer implicit information about mental states and causality, which has not been studied by most previous systems. 

Also related are commonsense knowledge bases 
\cite{OpenMindCommonsense,ConceptNet}. Our work complements these existing resources by providing commonsense relations that are relatively less populated in previous work.
For instance, ConceptNet contains only 25\% of our events, and only 12\% have relations that resemble intent and reaction.
We present a more detailed comparison with ConceptNet in Appendix~\ref{sec:conceptnetcomparison}.

\section{Conclusion}
We introduced a new corpus, task, and model for performing commonsense inference on textually-described everyday events, focusing on stereotypical intents and reactions of people involved in the events. Our corpus supports learning representations over a diverse range of events and reasoning about the likely intents and reactions of previously unseen events. We also demonstrate that such inference can help reveal 
implicit gender bias in movie scripts. 

\section*{Acknowledgments}
We thank the anonymous reviewers for their insightful comments. 
We also thank xlab members at the University of Washington, Martha Palmer, Tim O'Gorman, Susan Windisch Brown, Ghazaleh Kazeminejad as well as other members at the University of Colorado at Boulder for many helpful comments for our development of the annotation pipeline. 
This work was supported in part by
National Science Foundation Graduate Research Fellowship Program under
grant DGE-1256082, NSF grant IIS-1714566, 
and the DARPA CwC program through ARO (W911NF-15-1-0543).

\bibliographystyle{acl_natbib}
\bibliography{references}
\clearpage
\appendix
\section{Appendix}
\begin{figure*}[!htb]
    \centering
    \fbox{\includegraphics[width=\textwidth,trim=0 3cm 0 2cm]{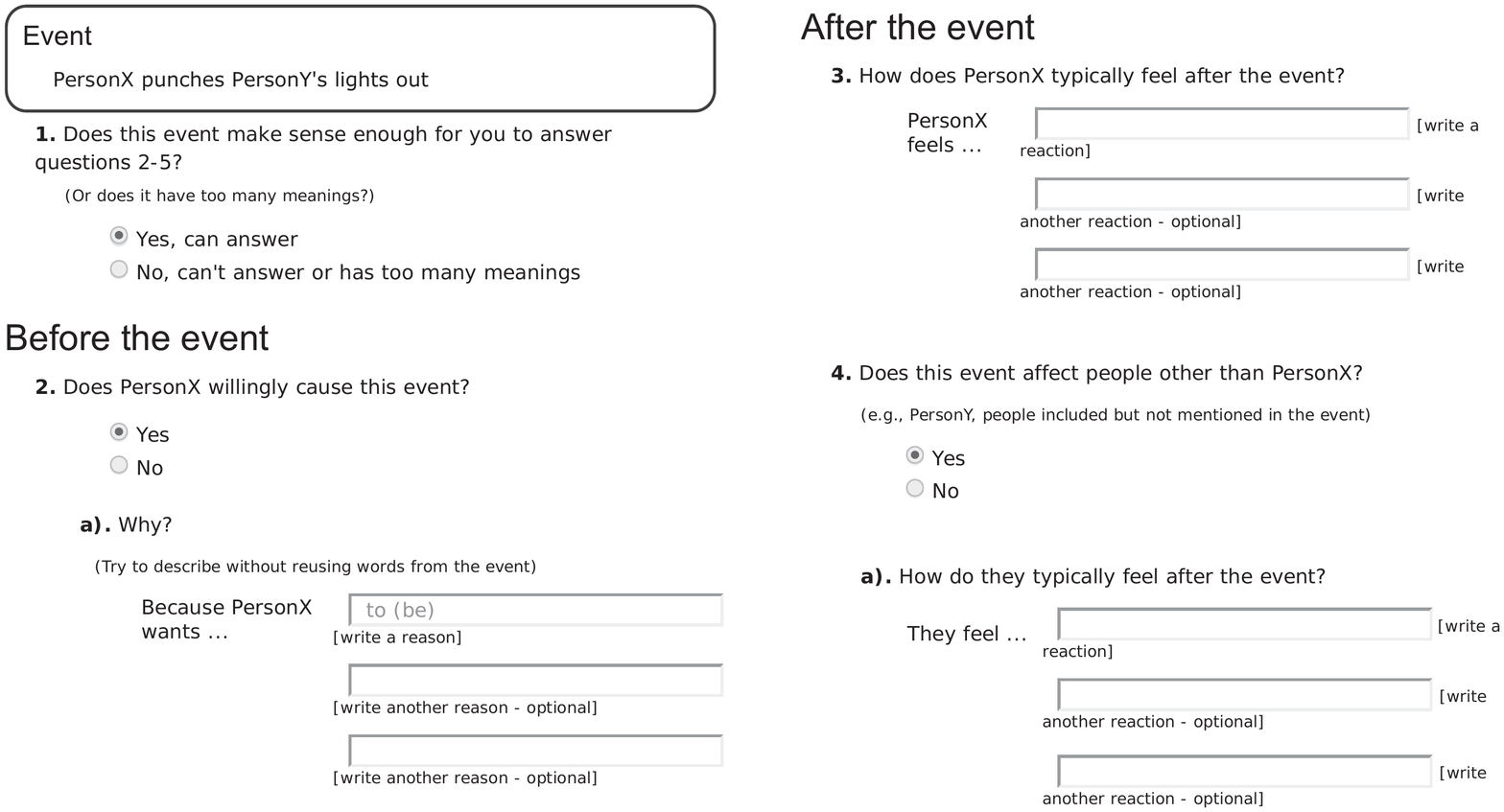}}
    \caption{Main event phrase annotation setup. Each event was annotated by three Amazon Mechanical Turk raters.
    \label{fig:mturkSetup}}
\end{figure*}

\subsection{Event Extraction}
\label{ss:event_extraction_more}
We balance the number of content words to ensure that the events are generalizable but still concrete enough to be labelled.
We only keep events with at least two and less than five content words, defined as words that are not stop words, person tags, or blanks.
We count phrasal verbs (such as ``get up'') as content word.
We limit the sets of events to those events that occur most frequently in our corpora, using corpus-specific thresholds.
\footnote{For ROC Story and Spinn3r events, we choose events with frequency at least five and 100, respectively.
For Syntactic Ngrams, we took the top 10000 events. 
}

\subsection{Annotation Setup}
Each event was presented to three different  raters recruited via Amazon Mechanical Turk. Raters were given the option to say that the event did not make sense (invalid), at which point they were not asked any other questions. If the rater marked the event as valid, they were required to answer the question about how PersonX typically feels after the event. Each  rater was paid $\$0.10$ per event. Additionally we annotated a small number of events where ``It'' was in the subject (e.g., \texttt{It rains all day}). For these events, we only asked  raters to say how other people typically feel after the event (if they marked the event as valid).

\section{Event2Mind Training Details}\label{sec:trainingdeets}
In our experiments, we use Adam to train for ten epochs, as implemented in Tensorflow \cite{tensorflow}. 

For baseline models, the dimension of the event encoded embedding is $H=300$.
For our BiRNN model, we also experimented with an embedding dimension of $H=100$.

We define the vocabulary as the tokens appearing in the training data events and annotations at least twice, plus the bigrams and trigrams that appear more than five times.  In cases where an annotation for the intent/reaction was left blank (because there was no intent or the event did not affect other people), we treated the annotation as equivalent to the word ``none''.  Because many of the annotations for intent started with ``to'' or ``to be'', we stripped these two words from the beginning of all intent annotations.

\section{Comparison with ConceptNet}
\label{sec:conceptnetcomparison}
We match our events with the event nodes in ConceptNet 
and find 6 ConceptNet relations that compare to our intent and reaction dimensions.
Specifically, we compare \textit{MotivatedByGoal, CausesDesire, HasFirstSubevent,} and \textit{HasSubevent} with the `XIntent' annotations, and `XReact' and `OReact' annotations with the \textit{Causes} and \textit{HasLastSubevent} relations. 
For each ConceptNet event, we then compute unigram overlap between our annotations and their ConceptNet proxy using the 6 relations.

We summarize overlap in Table~\ref{tab:conceptNet}, where we show that 75\% of Event2Mind events are not covered in ConceptNet.
We also show that while 12\% of our events have an edge with one of the 6 relations, the actual overlap between our annotations and the ConceptNet data is very low ($<$5\%). This overlap statistics indicates that our dataset provides new commonsense knowledge that is not covered by previous resources such as ConceptNet.

\begin{table}[]
\centering
\begin{tabular}{lc}\toprule
Overlap criterion & \multicolumn{1}{l}{\% of Event2Mind events} \\ \midrule
Any node & 25\% \\ \hline
\begin{tabular}[c]{@{}l@{}}All annotations,\\ with select relations\end{tabular} & 12 \% \\ \hline
\begin{tabular}[c]{@{}l@{}}XIntent, \\ with select relations\end{tabular} & 3\% \\\hline
\begin{tabular}[c]{@{}l@{}}XReact/OReact,\\ with select relations\end{tabular} & $<$1\%
\\\bottomrule
\end{tabular}
\caption{Event2Mind events overlap with ConceptNet events. While a non-trivial amount are represented in some capacity, few events have intent or reactions.} 
\label{tab:conceptNet}
\end{table}

\end{document}